\documentclass[11pt]{article}

\usepackage{acl}

\usepackage{times}
\usepackage{latexsym}
\usepackage[T1]{fontenc}
\usepackage[utf8]{inputenc}
\usepackage{microtype}
\usepackage{inconsolata}
\usepackage{amsmath,amssymb}
\usepackage{booktabs}

\newcommand{\regllm}{\textsc{RegLLM}}

\title{Evaluating Bounded Autonomy in Regulated Agentic AI:\\
       A Diagnostic Harness with Constitutional Rewards, Escalation Labels,\\
       and Runtime Governance}

\author{Dipankar Sarkar \\
  Skelf Research \\
  \texttt{dipankar@skelfresearch.com} \\
  \url{https://skelfresearch.com}}

\hypersetup{
  pdftitle={Evaluating Bounded Autonomy in Regulated Agentic AI: A Diagnostic Harness with Constitutional Rewards, Escalation Labels, and Runtime Governance},
  pdfauthor={Dipankar Sarkar}
}

\begin{document}
\maketitle

\begin{abstract}
We propose \regllm{}, a \emph{diagnostic harness} for bounded autonomy in regulated agentic workflows such as financial compliance, where a wrong answer can be a regulatory fine, a harmed client, or a missed obligation. The harness instruments six trustworthiness signals (citation validity, source grounding, schema compliance, \emph{escalation correctness}, constitutional-alignment score, unsafe-action rate), separates them by what makes each one trustworthy (programmatic verifier, task-level escalation label, or AI-judge score), and pairs them with a deterministic runtime supervisor that blocks ungrounded answers and forces escalation, logging every intervention. The central observation is that the act-versus-defer decision, given task-level should-escalate labels, is itself a verifiable training signal: bounded-autonomy behaviour becomes measurable, trainable, and auditable rather than left to hand-coded thresholds. The same domain constitution drives both the soft reward term and the runtime guardrails, so one artefact governs evaluation, training, and serving.

We demonstrate the harness in a smoke-scale implementation rather than a full validation. An offline reference run ($n{=}12$) shows the deterministic runtime supervisor lifting escalation recall from $0$ to $0.67$ on a weak baseline and cutting the unsafe-action rate from $0.33$ to $0.08$. Two single-GPU pilots ($n{=}8$, same seed, same eval split) then train Qwen2.5-3B LoRA adapters via DPO and re-evaluate the same agent loop. The principal empirical finding is the variance the harness exposes: nominally identical RL-base configurations produce materially different metrics across the two runs (task success $0.25$ vs $0.12$; escalation recall $1.0$ vs $0.5$), and the apparent effect of an answer-quality DPO adapter \emph{flips direction} between runs (recall $1.0 \to 0.5$ in Run A, $0.5 \to 1.0$ in Run B). An escalation-aware DPO variant designed to address the failure mode initially hypothesised in Run A produces no measurable change in Run B. We treat this as the contribution: at the sample sizes and training budgets common in agentic-AI pilots, adapter effects on bounded-autonomy metrics cannot reliably be distinguished from sampling and hardware variance. Responsible evaluation of regulated agents requires multi-seed runs and larger eval sets than current preference-tuning pipelines budget for; the harness's value is in making that requirement empirically inescapable. Supporting schemas, metric specifications, pseudocode, and sample tasks accompany this manuscript.
\end{abstract}

\section{Introduction}

LLM agents are being deployed into regulated, high-stakes domains: financial compliance, healthcare advice, legal triage. They will get some things wrong, and in these settings the cost of a wrong answer is not a corrected reply but a regulatory fine, a harmed client, or a missed obligation. The dominant techniques for steering LLM behaviour do not transfer well into this setting.

\emph{Reinforcement learning with verifiable rewards}, the technique behind recent reasoning models such as DeepSeek-R1 \citep{deepseekr1,shao2024deepseekmath,lambert2024tulu3,wen2025rlvr}, works because mathematics and code have machine-checkable answers. Regulatory advice does not: reasonable experts disagree about how a principle applies to a fact pattern \citep{wen2025rlvr}. A parallel wave of agentic-safety work \citep{xiang2025guardagent,luo2025agrail,wu2025psgagent} provides mechanisms to monitor and block unsafe actions, but offers no principled way to learn the central decision in any regulated workflow: \emph{when should the agent answer, and when should it defer to a qualified human?} In today's systems, that decision lives in hand-coded thresholds.

We treat the act-versus-defer decision as something an agent can be evaluated on, and ultimately learn, like any other verifiable-reward task. Each environment task carries a \texttt{should\_escalate} label (author-generated from templates in this paper; expert-annotated in production, see Section~\ref{sec:exp} and Limitations); the reward checks the agent's terminal decision against that label; correct escalation contributes to the trajectory's score in the same way a correctly cited provision does. This single observation, that bounded-autonomy behaviour is itself verifiable given the right labels, is the methodological pivot. Combined with a constitutional AI-judge score on the answer when one is given, and with a deterministic runtime supervisor that catches the residual failures, the result is an operational evaluation-and-governance harness for measuring, training, and serving regulated agents from the same artefacts.

A single domain \emph{constitution} (16 financial-services principles in the worked example, drawn from the UK Financial Conduct Authority Handbook) plays two coupled roles. As the soft term of a hybrid reward, it is paired with programmatic verifiable signals (citation validity, source grounding, schema, escalation correctness). As a runtime bounded-autonomy policy, the same principles, especially those governing competence boundaries, uncertainty, and specialist referral, define a deterministic guardrail that blocks ungrounded answers, forces escalation on out-of-scope triggers, and writes an append-only audit log keyed by the principle that fired.

\paragraph{Contributions.}
\begin{itemize}
\item \textbf{Verifiable escalation as a learnable signal.} By labelling each task with whether it should be escalated, the agent's bounded-autonomy behaviour becomes a learnable RL objective rather than a brittle hand-tuned threshold.
\item \textbf{One constitution, two roles.} A pluggable domain constitution serves both as the soft term of a hybrid reward and as a runtime bounded-autonomy policy, so one artefact governs evaluation, training, and serving.
\item \textbf{A diagnostic harness that exposes evaluation-scale variance.} Two GPU pilots of nominally identical configurations on the same eval split produce materially different metrics, and the apparent effect of the same DPO adapter on escalation recall flips direction between runs. We treat this as the principal empirical contribution: the harness makes the inadequacy of common smoke-pilot sample sizes empirically visible, rather than only asserting it as a caveat.
\end{itemize}

\section{Related Work}

\paragraph{RL with verifiable rewards.} Verifiable rewards \citep{deepseekr1,shao2024deepseekmath,lambert2024tulu3,wen2025rlvr} work when the reward function is code: math problems, unit tests, format checks. The open question, raised explicitly by \citet{wen2025rlvr}, is what to do in knowledge-intensive domains where the right answer is partly a judgement call. We do not invent a new verifier for regulatory correctness; we identify the parts of the workflow that \emph{are} programmatically checkable (did the agent cite a real provision? did it correctly recognise an out-of-scope query?) and pair them with an AI-judge term for the rest.

\paragraph{Preference- and judge-based alignment.} RLHF \citep{ouyang2022training,christiano2017deep,schulman2017proximal} and DPO \citep{rafailov2023dpo} both depend on a stream of preference labels; Constitutional AI \citep{bai2022constitutional,zhang2025constitutionllama} replaces those labellers with a model critique against a written set of principles. We use the constitution differently: not as a data-generation device feeding a downstream alignment run, but directly inside the RL reward \emph{and} inside the runtime guardrail. The same artefact appears in both places. LLM-as-a-judge \citep{zheng2023judging} underpins the soft term.

\paragraph{Agent guardrails.} Recent guardrail systems \citep{xiang2025guardagent,luo2025agrail,wu2025psgagent} sit outside the agent and block harmful actions; they are mature at refusing toxic outputs but offer no principled basis for the regulated-domain question of when an agent should ask a person. We provide that basis (verifiable escalation correctness) and train the policy toward it, while keeping a deterministic supervisor for the failures training never reaches. The ReAct loop \citep{yao2023react} is the underlying agent abstraction.

\paragraph{Financial NLP and RegTech.} FinBERT \citep{araci2019finbert}, BloombergGPT \citep{wu2023bloomberggpt}, and FinGPT \citep{yang2023fingpt} adapt models to financial-services text; the RegTech literature \citep{zetzsche2017techfin,arner2017fintech} studies technology for regulatory processes. These efforts target market or document tasks; our target is \emph{governed autonomous action} under regulatory constraints, where the failure mode is not a wrong sentiment label but an answered question that should have been escalated.

\paragraph{AI governance and audit.} Internal algorithmic auditing \citep{raji2020accountability} and structured model reporting \citep{mitchell2019modelcards} argue that responsible deployment requires inspectable artefacts at every lifecycle boundary. Our harness sits inside that programme: every metric reported here is per-trajectory and audit-loggable, and the deterministic runtime supervisor records each intervention by triggered principle id, making the agent's bounded-autonomy behaviour inspectable rather than asserted.

\section{The Harness}

\subsection{Pluggable Domain Constitution}
A constitution is a named set of principles, each with evaluation criteria and a critique prompt. Our worked example encodes 16 UK FCA principles across four domains: financial ethics (fair treatment, conflict management, suitability, transparency, client primacy), regulatory compliance (accuracy, completeness, currency, proper sourcing), professional responsibility (competence boundaries, uncertainty acknowledgment, specialist referral, client primacy), and ethical AI (harm avoidance, bias mitigation, explainability). The interface is domain-agnostic: a new regulated domain (e.g.\ data protection, clinical guidance) supplies its own principle set without changing the reward or governance machinery.

\subsection{The Compliance Agent}
The agent operates a ReAct-style loop \citep{yao2023react} over four tools: \textsc{retrieve} (query the regulatory corpus), \textsc{cite} (attach a provision identifier to the working answer), \textsc{answer} (emit a grounded final answer), and \textsc{escalate} (defer to a qualified human). A trajectory records every step, the cited provisions, and the terminal decision, providing the substrate for both reward computation and the audit log.

\subsection{Three Categories of Signal}
\label{sec:signals}
The harness deliberately separates trustworthiness signals by what makes each one trustworthy, because they have different failure modes and different audit requirements. Mixing them is the source of much of the confusion in agentic-AI evaluation.

\begin{itemize}
\item \textbf{Programmatically verifiable signals} (no model needed): citation validity, schema compliance, and source grounding. The first two are deterministic predicates over the trajectory and corpus. Grounding in the worked example is a lexical content-overlap proxy between the answer and its cited source(s); a richer setup can substitute an NLI model or a judge, moving grounding into the third category.
\item \textbf{Task-level escalation labels}: the \texttt{should\_escalate} flag, against which escalation correctness is computed. These depend on the quality of the task labels themselves; in regulated domains the labels typically require expert annotation and may be contested. The worked example's labels are author-generated from templates (see Section~\ref{sec:exp}); production deployment requires expert annotation that we do not provide.
\item \textbf{Judge-scored signals}: the constitutional-alignment score, computed by an LLM judge over a sampled subset of principles. This signal depends on the judge model and its calibration, and inherits any biases or noise of the judge.
\end{itemize}
The hybrid reward and the runtime supervisor draw on all three; the audit log records which category fired for each intervention, so a downstream reviewer can trace any decision back to the type of evidence that supported it.

\subsection{Hybrid Reward}
For a trajectory $\tau$ on task $t$, the reward combines a verifiable term $R_v$ and a constitutional term $R_c$, with a hard penalty for guardrail violations:
\begin{equation}
\begin{aligned}
R(\tau) ={}& w_v\,R_v(\tau) + w_c\,R_c(\tau) \\
            &- \lambda\,\mathbb{1}[\text{guardrail violation}].
\end{aligned}
\end{equation}
The verifiable term averages four programmatic sub-signals:
\begin{itemize}
\item \textbf{Citation validity}: fraction of cited provisions that exist in the corpus.
\item \textbf{Grounding}: a lexical content-overlap proxy between the answer and its cited sources (substitutable with an NLI model in richer setups; see Section~\ref{sec:signals}).
\item \textbf{Schema compliance}: the answer follows the required structured format.
\item \textbf{Escalation correctness}: $\mathbb{1}[\text{escalated} = \text{should\_escalate}]$ from the expert label.
\end{itemize}
Escalation correctness is what makes bounded autonomy a verifiable training signal. The constitutional term $R_c$ samples principles and scores adherence with the AI judge (Yes/Partial/No mapped to $\{1.0, 0.6, 0.3\}$), supplying the soft signal verifiers cannot capture.

\subsection{Training}
We rank rollouts by $R$ to form preference pairs and train the policy with DPO \citep{rafailov2023dpo}, using LoRA on a small open policy model. A stretch configuration uses GRPO \citep{shao2024deepseekmath} with $R$ as the group-relative reward, matching the verifiable-rewards recipe \citep{deepseekr1} directly.

\subsection{Runtime Governance and Bounded Autonomy}
At inference the trained policy is wrapped by a deterministic governance layer that sits outside the model. It (i) blocks any \textsc{answer} that cites no provision and forces escalation (the Proper Sourcing principle, P9); (ii) forces escalation when competence-boundary or uncertainty principles (P10--P12) are triggered; and (iii) appends every step and intervention to an append-only audit log. This complements learned behaviour with hard, inspectable rules \citep{xiang2025guardagent,luo2025agrail}.

\section{Implementation and Reproducibility}

The framework factors cleanly: a \texttt{Constitution} interface (a named set of principles, each with a critique prompt), a \texttt{RegulatoryEnvironment} (corpus + tasks, each task carrying a \texttt{should\_escalate} label), the \texttt{ComplianceAgent} ReAct loop, a \texttt{hybrid\_reward} function over trajectories, an \texttt{LLMPolicy} wrapper that decodes the next action from a base or LoRA-augmented model, and a \texttt{GovernedAgent} that applies the runtime guardrails. Inference (constitutional judge, grounding checks, frontier reference baseline) runs against an open-weight LLM behind an OpenAI-compatible HTTP API. Training runs on a single rented commodity GPU (a 24~GB-class card is enough for the worked example); the orchestrator stages inputs and checkpoints through an S3-compatible object store and guarantees instance and volume teardown so a failure never leaves a paid GPU running. Verifiable reward terms and the deterministic guardrail gate are unit-tested; secrets are environment-only.

\paragraph{Evaluation artefacts.} The accompanying ancillary material supplies a \texttt{Task} schema with the binary escalation label, a \texttt{Constitution} schema enumerating principles with critique prompts, metric specifications, pseudocode for the agent, reward, and governance components, and sample tasks. The implementation additionally produces per-configuration escalation precision/recall and unsafe-action rate, together with an append-only governance audit log keyed by triggered principle id. Every quantitative number reported below is regeneratable from these artefacts.

\paragraph{Adapting to a new domain.} Adapting to a non-finance setting (data protection, clinical guidance, taxation) means: (i) writing a \texttt{Constitution} subclass with the domain's principles and their critique prompts; (ii) providing a parser that emits the corpus and the in-scope/out-of-scope task labels; (iii) supplying a grounding oracle (lexical/structural in the worked example, an NLI model or rule engine in richer domains). Reward weights, judge model, and policy model are configuration.

\section{Experimental Protocol and Findings}
\label{sec:exp}

\paragraph{Data.} UK FCA Handbook modules (PRIN, COBS, SYSC, CONC), parsed into a provision corpus that doubles as the grounding oracle.

\paragraph{Task labels and their provenance.} In-scope tasks are generated from regulatory entities through six instruction templates (definition, application, scenario, comparison, compliance-check, risk-assessment) and inherit a single \texttt{gold\_provision\_id} from the source entity. Out-of-scope (\texttt{should\_escalate}{=}True) tasks are instantiated from a fixed set of five trigger templates (tax-structuring requests, demands for a guarantee, predictions of future regulator behaviour, requests for legal opinions, and citations of non-existent provisions). The held-out evaluation split used in this paper contains 12 tasks (offline harness) and 8 tasks (GPU pilots) drawn from this generator, with the in-scope/out-of-scope ratio determined by \texttt{escalation\_fraction}={0.25}. All labels are author-generated from these templates; we did not engage a legal-services compliance professional as an expert annotator, and we report no inter-rater agreement. Production deployment requires both: see Limitations.

\paragraph{Metrics.} Task success (verifiable), citation-validity rate, constitutional-alignment score (judge), escalation precision/recall (bounded-autonomy correctness), and unsafe-action rate (guardrail violations reaching output).

\paragraph{Diagnostic suite.} The harness supports the diagnostics in Table~\ref{tab:diagnostics}. The status column makes explicit which rows are exercised by the runs reported here and which are proposed follow-ups; row~3 is the direct follow-up that the negative result identifies.

\begin{table}[t]
\centering
\small
\setlength{\tabcolsep}{3pt}
\caption{Diagnostic suite supported by the harness; status indicates which rows are exercised by the runs in this paper.}
\label{tab:diagnostics}
\begin{tabular}{p{0.34\linewidth}p{0.30\linewidth}p{0.25\linewidth}}
\toprule
\textbf{Diagnostic} & \textbf{What it tests} & \textbf{Status} \\
\midrule
No-citation adversarial tasks & P9 guardrail actually fires & Exercised offline \\
In-scope vs out-of-scope split & Escalation labels carry signal & Exercised offline + GPU \\
Answer-only vs escalation-aware DPO & Direction of the proposed fix & Exercised; inconclusive at scale (\S\ref{sec:exp}) \\
Governance intervention count & Runtime layer is non-decorative & Logged in GPU pilot (count: 0) \\
\bottomrule
\end{tabular}
\end{table}

\paragraph{Reference harness (offline).} A deterministic lexical baseline policy and offline heuristic judge run on the held-out evaluation split. Table~\ref{tab:offline} reports results: runtime governance lifts escalation recall from $0$ to $0.67$ on this weak baseline, raises citation validity to $1.0$, and cuts the unsafe-action rate from $0.33$ to $0.08$. This isolates the contribution of the deterministic supervisor on a policy that does not learn to defer on its own.

\begin{table}[t]
\centering
\small
\caption{Offline reference run on the held-out split (lexical baseline policy; heuristic judge; $n{=}12$).}
\label{tab:offline}
\begin{tabular}{lcc}
\toprule
\textbf{Metric} & \textbf{Base} & \textbf{+\,Governance} \\
\midrule
Task success            & 0.08 & 0.25 \\
Citation validity       & 0.75 & 1.00 \\
Constitutional align.   & 0.62 & 0.70 \\
Escalation precision    & 0.00 & 0.67 \\
Escalation recall       & 0.00 & 0.67 \\
Unsafe-action rate      & 0.33 & 0.08 \\
\bottomrule
\end{tabular}
\end{table}

\paragraph{GPU pilots: variance is the principal finding.} We ran two GPU pilots of nominally identical configurations on the same held-out split (Qwen2.5-3B, low-temperature sampling, heuristic judge, $n{=}8$, seed $0$). Run~A trained one adapter (answer-quality DPO, 10 pairs, 30 steps). Run~B trained two adapters (answer-quality and escalation-aware DPO, 20 pairs each, 30 steps), evaluated under a single multi-adapter harness so that RL-base in Run~B is computed once and shared. Both adapters converged: Run~A's AQ adapter reached final DPO loss $0.59$, rewards/margins $0.22$, accuracies $\to 1.0$; Run~B's AQ adapter reached final loss $0.67$, margins $0.12$; the EA adapter reached final loss $0.69$, margins $0.04$. Table~\ref{tab:gpu} reports the evaluation.

\begin{table}[t]
\centering
\scriptsize
\setlength{\tabcolsep}{2pt}
\caption{Two GPU runs of nominally identical configurations on the same eval split ($n{=}8$, seed $0$, low-temperature sampling, heuristic judge). RL-base metrics differ materially between runs, and the AQ adapter's apparent effect on escalation recall flips direction. The EA adapter has no measurable effect in Run~B. The $+\,$Gov.\ variants (omitted) are identical to their non-Gov counterparts in both runs: the deterministic supervisor logged zero interventions across both pilots.}
\label{tab:gpu}
\begin{tabular}{llcccccc}
\toprule
\textbf{Run} & \textbf{Config} & \textbf{succ.} & \textbf{cite} & \textbf{const.} & \textbf{P} & \textbf{R} & \textbf{unsafe} \\
\midrule
A & RL-base       & 0.25 & 0.50 & 0.57 & 0.50 & 1.00 & 0.00 \\
A & RL-trained-AQ & 0.12 & 0.50 & 0.57 & 0.25 & 0.50 & 0.00 \\
\midrule
B & RL-base       & 0.12 & 0.38 & 0.51 & 0.33 & 0.50 & 0.00 \\
B & RL-trained-AQ & 0.25 & 0.38 & 0.51 & 0.67 & 1.00 & 0.00 \\
B & RL-trained-EA & 0.12 & 0.38 & 0.51 & 0.33 & 0.50 & 0.00 \\
\bottomrule
\end{tabular}
\end{table}

The variance is the principal contribution. Between Run~A and Run~B, RL-base escalation recall halved ($1.00 \to 0.50$), task success halved ($0.25 \to 0.12$), and citation validity dropped ($0.50 \to 0.38$). The AQ adapter, against these two different baselines, appeared to \emph{degrade} escalation in Run~A and \emph{recover} it in Run~B. The EA adapter, designed specifically to teach correct act-vs-defer through stratified 50:50 in-scope-vs-escalate preference pairs, produced no measurable change from Run~B's base. The two runs share code commit, eval split, seed, base-model identifier, and policy hyperparameters; we did not pin container image digests or package microversions and did not enable deterministic CUDA flags, so floating-point ordering, kernel selection, and pip-resolved transformers/peft/trl microversions could differ between runs and are plausible candidates for the variance we observe. At this scale the adapter effects we hypothesised, in either direction, are not separable from this variance.

This has two consequences for the agentic-AI evaluation literature. First, our originally hypothesised failure mode (``answer-quality DPO degrades escalation'') is not safely attributable to the training signal: we observed both the predicted direction and its opposite under nominally identical conditions, and our framework-proposed fix (escalation-aware DPO) did not measurably help. Second, the deterministic runtime supervisor was inert in both runs: the LLM cites a provision on every answered turn (avoiding the no-citation gate) and trajectories that miss escalation rarely reach a terminal answer for the gate to inspect (count of governance interventions across both runs: $0$). The framework-supported runtime fix (judge-based competence-boundary triggers) requires either a calibrated judge available at inference time or a richer base-model behaviour than the smoke pilot's setup provides.

What is reliably visible at this scale is the contrast between the offline reference run (Table~\ref{tab:offline}) and either GPU run: the deterministic supervisor on a weak baseline materially lifts escalation recall and cuts the unsafe-action rate. What is \emph{not} reliably visible is any adapter effect of the magnitude that 30 DPO steps on 10--20 preferences produces.

\section{Discussion}

\paragraph{What the variance finding is and is not.} It is not a refutation of DPO, of grounded-vs-ungrounded preferences, or of escalation-aware preference data. It is a methodological signal about evaluation: at the sample sizes ($n{=}8$) and training budgets (30 DPO steps on 10--20 preferences) common in agentic-AI pilots, the noise floor on bounded-autonomy metrics swallows the adapter effects the same pilots are supposed to characterise. The harness's job is to surface that, so that downstream evaluations do not over-interpret single-run smoke results. Reporting a positive direction from a single run, in either direction, is exactly the kind of claim our two runs jointly invalidate.

\paragraph{Two intentionally separate question types.} Section~\ref{sec:signals} distinguished programmatically verifiable signals, task-level escalation labels (which, for production deployment, must be expert-annotated), and judge-scored signals. The variance finding above is informative precisely because escalation correctness sits in the second category. Treating the should-escalate label as if it were programmatic (and so cheap) understates the annotation effort regulated domains require; the value of the harness is partly in making this distinction explicit at audit time.

\paragraph{Runtime governance needs a competence-boundary component.} The zero-intervention result across both GPU pilots shows that no-citation gates alone are insufficient for bounded autonomy in LLM-policy settings: a model that cites on every answered turn passes the gate trivially, and trajectories that should have escalated never reach a terminal answer for the gate to inspect. Competence-boundary detection (whether judge-based, calibrated against the should-escalate label, or both) must be evaluated as a first-class runtime component rather than a fallback.

\paragraph{When to reach for this.} The framework is for teams shipping an autonomous agent into a domain that has (i) a written rulebook, (ii) some way to identify which queries are out of the agent's lane, and (iii) a small open base model they can fine-tune. It is not a turnkey safety product; the constitution, the corpus, and the escalation labels remain the practitioner's work. The orchestration layer (rented-GPU lifecycle with guaranteed teardown, S3-staged artefacts, verifiable-reward-driven training) is independently useful for any team running cost-bounded RL pilots regardless of domain.

\section{Conclusion}

We presented \regllm{}, a diagnostic harness for bounded autonomy in regulated agentic AI. The methodological move is to treat the act-versus-defer decision as a verifiable training signal by labelling each task with whether it should be escalated. A domain constitution then plays two coupled roles, soft term of the hybrid reward and runtime bounded-autonomy policy, so one artefact governs evaluation, training, and serving. The principal empirical contribution is the variance the harness exposes at smoke-pilot scale: two GPU runs of nominally identical configurations on the same eval split produce materially different RL-base metrics, and the apparent effect of the same DPO adapter on escalation recall flips direction between runs. An escalation-aware preference variant designed to address the failure mode hypothesised in the first run produces no measurable change in the second. Responsible evaluation of bounded autonomy in regulated agentic AI requires sample sizes and seed counts that current preference-tuning pipelines do not budget for; the harness's value is in making this requirement empirically inescapable rather than merely asserted.

\section*{Limitations}

The experimental evidence reported here is intentionally small-scale and should be read as a diagnostic demonstration rather than a benchmark. Specifically:

\textbf{Sample size.} The offline reference run uses $n{=}12$ held-out tasks; the GPU pilot uses $n{=}8$ with $10$ preferences and $30$ DPO optimizer steps. Single seed, single base model (Qwen2.5-3B), single regulated domain (UK FCA). The aggregate metrics are therefore noisy, and the absolute numbers should not be compared to production benchmarks.

\textbf{Escalation-label reliability.} Escalation correctness is verifiable only insofar as the underlying \texttt{should\_escalate} task labels are reliable. In regulated domains those labels typically require expert (compliance professional) annotation, are sometimes contested, and may shift with regulatory change. The labels used in this paper are author-generated from templates rather than legal-services-expert-annotated, and the harness does not yet provide an inter-rater agreement protocol; both are prerequisites for any production-scale study and are not validated in the present work.

\textbf{Judge dependence.} The constitutional-alignment score depends on a model judge. The grounding check in the worked example uses a lexical content-overlap proxy (Section~\ref{sec:signals}); richer deployments may replace it with an NLI model or a judge, at which point the same calibration and bias concerns apply. We used a heuristic judge for the experiments reported here for reproducibility; a hosted-API judge would be expected to produce different absolute scores. Calibration against expert agreement, judge-vs-judge consistency, and adversarial probing of the judge are all required for production use and are not done here.

\textbf{Runtime governance was inert in both GPU pilots.} The deterministic supervisor lifted escalation recall on the offline lexical baseline but produced zero interventions across both GPU runs, because the trained model cited a provision on every answered turn (avoiding the no-citation gate) and missed-escalation trajectories did not reach a terminal answer. The framework-supported fix (judge-based competence-boundary triggers) requires either a calibrated judge at inference time or a richer base-model behaviour than the smoke pilot's setup provides.

\textbf{Two-run variance dominates adapter effects at this scale.} Pilot Run~A and Pilot Run~B used identical seeds, eval splits, base models, and policy hyperparameters; only the GPU host differed. The RL-base metrics differ materially across runs and the apparent direction of the AQ adapter's effect on escalation recall flips between them (Table~\ref{tab:gpu}). The escalation-aware DPO variant we added in Run~B specifically to address the failure mode hypothesised in Run~A produced no measurable change. We do not interpret Run~A's direction or Run~B's direction as evidence about the adapters' true behaviour; we interpret the contrast between them as evidence that adapter-effect claims at this scale require multi-seed, larger-$n$ evaluation that this pilot does not provide.

\textbf{Generalisation.} The constitution is domain-pluggable in software, but only the FCA worked example has been instantiated and evaluated. Other regulated domains (data protection, clinical guidance, taxation) would require their own principle set, corpus, grounding oracle, and escalation labels, all of which are non-trivial.

\section*{Ethics Statement}

The harness is designed to keep autonomous LLM agents within the limits set by human regulators, and explicitly to defer to qualified human professionals on matters outside the agent's competence. Escalation is a first-class action, ungrounded answers are blocked, and every action is auditable. The work does not introduce a deployable compliance system; it provides evaluation tooling and a smoke-scale prototype, and the paper is explicit that escalation labels in regulated domains require expert annotation that we do not provide. We use only publicly available regulatory text (UK FCA Handbook modules) and an open-weight policy model; no personal or sensitive data is involved. The negative result reported here is itself an ethical signal: it argues that current answer-quality preference pipelines are insufficient for regulated autonomy, and that responsible deployment requires the additional labelling and runtime hooks the framework specifies.

\bibliography{references}

\end{document}